\documentclass{article}
\usepackage[final]{neurips_2023}
\usepackage{hyperref}
\usepackage{natbib}
\setcitestyle{numbers,square}
\usepackage[utf8]{inputenc} 
\usepackage[T1]{fontenc}    
\usepackage{hyperref}       
\usepackage{url}            
\usepackage{booktabs}       
\usepackage{amsfonts}       
\usepackage{nicefrac}       
\usepackage{microtype}      
\usepackage{xcolor}         
\usepackage{multirow}
\usepackage{graphicx}
\usepackage[flushleft]{threeparttable}
\usepackage{adjustbox}
\usepackage{subcaption}

\title{WanJuan-CC: A Safe and High-Quality Open-sourced English Webtext Dataset}

\author{%
Jiantao Qiu\thanks{Co-first authors.} \quad Haijun Lv\footnotemark[1] \quad Zhenjiang Jin \quad Rui Wang \quad Wenchang Ning\\
\textbf{Jia Yu} \quad \textbf{Chaobin Zhang} \quad \textbf{Zhenxiang Li} \quad \textbf{Pei Chu} \quad \textbf{Yuan Qu} \quad \textbf{Jin Shi} \\
\textbf{Lindong Lu} \quad \textbf{Runyu Peng} \quad \textbf{Zhiyuan Zeng} \quad \textbf{Huanze Tang} \quad \textbf{Zhikai Lei} \\
\textbf{Jiawei Hong} \quad \textbf{Keyu Chen} \quad \textbf{Zhaoye Fei} \quad \textbf{Ruiliang Xu} \quad \textbf{Wei Li} \\
\textbf{Zhongying Tu} \quad \textbf{Lin Dahua}\thanks{Co-corresponding author} \quad \textbf{Yu Qiao}\footnotemark[2] \quad \textbf{Hang Yan}\footnotemark[2] \quad  \textbf{Conghui He}\footnotemark[2] \\
Shanghai Artifcial Intelligence Laboratory\\
Shanghai, 200232, China\\
\texttt{ \{qiaoyu, lindahua, yanhang, heconghui\}@pjlab.org.cn} \\
}

\begin{document}

\maketitle

\begin{abstract}
This paper presents WanJuan-CC, a safe and high-quality open-sourced English webtext dataset derived from Common Crawl data. The study addresses the challenges of constructing large-scale pre-training datasets for language models, which require vast amounts of high-quality data. A comprehensive process was designed to handle Common Crawl data, including extraction, heuristic rule filtering, fuzzy deduplication, content safety filtering, and data quality filtering. From approximately 68 billion original English documents, we obtained 2.22T Tokens of safe data and selected 1.0T Tokens of high-quality data as part of WanJuan-CC. We have open-sourced 100B Tokens from this dataset. The paper also provides statistical information related to data quality, enabling users to select appropriate data according to their needs. To evaluate the quality and utility of the dataset, we trained 1B-parameter and 3B-parameter models using WanJuan-CC and another dataset, RefinedWeb. Results show that WanJuan-CC performs better on validation datasets and downstream tasks.
\end{abstract}

\section{Introduction}
Recently, large-scale language models have achieved significant accomplishments in numerous natural language processing tasks. Notable research outcomes such as GPT3\cite{DBLP:conf/nips/BrownMRSKDNSSAA20}, BLOOM\cite{DBLP:journals/corr/abs-2211-05100}, LLAMA\cite{DBLP:journals/corr/abs-2302-13971}, Palm\cite{DBLP:journals/jmlr/ChowdheryNDBMRBCSGSSTMRBTSPRDHPBAI23}, Falcon\cite{DBLP:journals/corr/abs-2306-01116}, GPT4\cite{DBLP:journals/corr/abs-2303-08774}, InternLM\cite{2023internlm}, ChatGLM\cite{zhang2023chatglm}, and Baichuan\cite{DBLP:journals/corr/abs-2309-10305} have been reported successively. These language models, pre-trained on large-scale data, demonstrate excellent performance across various NLP tasks, even without the need for fine-tuning with labeled data\cite{DBLP:conf/nips/BrownMRSKDNSSAA20}.

Existing research indicates that there exists a scaling relationship, known as the \textbf{Scaling Law}, among optimal model parameters, training data volume, and total computational overhead in the training process of large models\cite{DBLP:journals/corr/abs-2203-15556}. This implies that to train more powerful models, we need to simultaneously increase the number of model parameters and the scale of the pre-training dataset. Studies have shown that training a language model with 175B parameters requires approximately 3.7T tokens of high-quality pre-training data. However, traditional data schemes that collect and custom clean from specific data sources can no longer meet this scale of data demand, posing new challenges for the construction of pre-training datasets.

Webtext, text data scraped and extracted from the internet, is a key resource for building pre-training datasets. Many large language model studies \cite{DBLP:conf/nips/BrownMRSKDNSSAA20,DBLP:journals/corr/abs-2211-05100,DBLP:journals/corr/abs-2302-13971,DBLP:journals/jmlr/ChowdheryNDBMRBCSGSSTMRBTSPRDHPBAI23} have claimed to incorporate webtext data as part of their pre-training data, mixing it with other high-quality data sources for model training. Some research even demonstrates that using appropriately processed web-scraped data alone can achieve results similar to those obtained with mixed datasets of high-quality data \cite{DBLP:journals/corr/abs-2306-01116}.

Common Crawl\cite{CommonCrawl} maintains a freely accessible web-crawled database, significantly reducing the cost of re-crawling data. It provides historical web page data since 2008, making it an essential source for webtext data. However, the quality of data in Common Crawl is inconsistent, with a substantial amount of low-quality data such as format errors, duplicate content, and advertising information, which are not beneficial for model training. Removing these low-quality data can optimize the training effect of the model. Moreover, Common Crawl contains a large amount of unsafe content, such as toxic and pornographic materials, which could lead to unfriendly outputs from the model. The presence of personally identifiable information (PII) in the data could also infringe on user privacy. Therefore, to enhance the safety of the model, it is necessary to filter or mask these data.

To address the aforementioned issues, we designed and implemented a process for handling Common Crawl data, which includes data extraction, heuristic rule filtering, fuzzy deduplication, content safety filtering, and data quality filtering. From the Common Crawl, we selected and extracted approximately 68B original English documents. After undergoing this series of processing steps, we obtained 2.22T Tokens of safe data. Further quality filtering yielded 1T Tokens of high-quality data, forming what we refer to as the WanJuan-CC dataset. We have extracted 100B Tokens from WanJuan-CC as open-source data, accompanied by some statistical information related to data quality, allowing users to select suitable data based on their specific needs.

In summary, the main contributions of this study are as follows:
\begin{enumerate}
\item We designed and implemented a process for handling Common Crawl data, which includes steps such as data extraction, heuristic rule filtering, fuzzy deduplication, content safety filtering, and data quality filtering.
\item We extracted raw data from approximately 68 billion documents in Common Crawl. After processing through the aforementioned steps, we obtained 2.22T Tokens (9.71TB) of safe data, and selected 1.0T Tokens (4.45TB) of high-quality data as part of WanJuan-CC. We have open-sourced 100B Tokens from this dataset.
\item We included some statistical information related to data quality in the dataset, allowing users to select appropriate data according to their needs.
\item We used WanJuan-CC and RefinedWeb as training data to train 1B-parameter and 3B-parameter models, using perplexity on the validation dataset and accuracy on downstream tasks as evaluation metrics. The results show that the data quality of WanJuan-CC is superior to that of RefinedWeb.
\end{enumerate}
\section{Related Work}

\subsection{Webtext Datasets}

By openly sharing meticulously processed datasets with the community, we can propel advancements in the field of large language models and reduce the cost of data processing for various research institutions. Currently, some publicly available datasets directly utilize the Common Crawl database or adopt the OSCAR dataset \cite{suarez2019asynchronous} based on Common Crawl. This primarily includes studies such as RefinedWeb \cite{DBLP:journals/corr/abs-2306-01116}, Redpajama \cite{together2023redpajama}, and Dolma \cite{soldaini2023dolma}.

\begin{table}[htbp]
    \centerline{
    \centering
    \begin{adjustbox}{width=15cm}
    \begin{threeparttable}
    \caption{Comparison of Pre-training Datasets Including Common Crawl Data}
    
    \begin{tabular}{c|c|c|c|c|c|c}
    \toprule
    Dataset    &  Total Size\tnote{$\star$} & CC Dumps & Quality Cls\tnote{$\dagger$} & URL/Words    & Toxic/Porn Cls\tnote{$\dagger$} & PII Mask     \\
    \midrule
    Redpajama  & 1.2T                & 46       & \color{red}\textbf{Yes}      & No           & No                              & No           \\ 
    Dolma      &  3.1T                & 24       & No                           & No           & Toxic                           & \color{red}\textbf{Yes}  \\ 
    RefinedWeb &  5.0T                & 94       & No                           & \color{red}\textbf{Yes}  & No                              & No           \\ 
    WanJuan-CC    & 1.0T                & 90       & \color{red}\textbf{Yes}      & \color{red}\textbf{Yes}  & \color{red}\textbf{Both}                    & \color{red}\textbf{Yes}  \\ 
    \bottomrule
    \end{tabular}
    \begin{tablenotes}
        \item[$\star$]: Count the total size of the dataset claimed by the publisher.
        \item[$\dagger$]: Quality Cls and Toxic/Porn Cls mean use model-based classifier to filter the data.

    \end{tablenotes}
    \label{tab:related_works}
    \end{threeparttable}
    \end{adjustbox}
    }
\end{table}

The primary attributes of datasets processed based on Common Crawl include dataset size, the number of CC dumps, as well as the safety of the dataset, personal privacy protection, and data quality filtering methods. These are widely recognized features. We have conducted relevant attribute statistics for the aforementioned datasets and the dataset in this study, and the results are summarized in Table \ref{tab:related_works}. According to the statistical data in the table, among the existing public datasets, only RefinedWeb covers data from more than 90 CC dumps. Currently, no public dataset fully covers content safety measures in terms of toxicity, pornography, and personally identifiable information (PII). Moreover, only redpajama has adopted a model-based quality filtering method, while other datasets merely use heuristic rules for quality filtering. The open-sourced dataset in this study covers 90 CC dumps. In addition to keyword and URL-based blocking, we also used a model-based approach to exclude data containing toxic and pornographic content, and regular expressions were utilized to mask PII information. Finally, we adopted a model-based quality filtering method to screen out relatively high-quality data. Ultimately, our open-sourced dataset includes 1TB of high-quality data.

\subsection{Data Utility}
While Common Crawl provides a vast and diverse set of data, it is also interspersed with a considerable amount of low-quality information. This content could negatively impact the performance of models and may lead to the generation of unsafe outputs, thereby reducing the practicality of the models. We collectively refer to these two issues at the data level as \textit{Data Safety} and \textit{Data Quality}, which are overall referred to as \textit{Data Utility}. Herein, we summarize some methods for enhancing \textit{Data Utility} identified in related research.

\subsubsection{Data Safety}
Employing keywords and domain names to filter certain data is a fundamental measure for data security. For instance, RefinedWeb\cite{DBLP:journals/corr/abs-2306-01116} utilizes a blacklist comprising 4.6 million URLs to filter potentially harmful web pages, and scores each URL based on an unsafe word list, setting thresholds to screen the data. Similarly, Roots\cite{DBLP:conf/nips/LaurenconSWAMSW22} also adopts a filtering method based on the density of sensitive words.

Several studies have employed model-based methods to filter out toxic and pornographic data. Toxic content is generally defined as material that causes users to leave their current reading or discussion, posing a significant problem for the primary applications of large language models. Pornography, on the other hand, is typically defined as data containing explicit sexual content. This type of data constitutes a considerable proportion in Common Crawl, not only reducing the ratio of useful data but also potentially leading to unsafe outputs from the model. In Dolma\cite{soldaini2023dolma}, a toxic classifier trained on the Jigsaw dataset\cite{jigsawtoxic} was used to filter the data.

The Common Crawl dataset contains a significant number of pages with Personally Identifiable Information (PII), and training models on this data can lead to privacy breaches. To address this issue, the current common practice is to use regular expressions to mask such information, as demonstrated in works like Dolma\cite{soldaini2023dolma}, Roots\cite{DBLP:conf/nips/LaurenconSWAMSW22}, and Wudao\cite{DBLP:journals/aiopen/YuanZDDLCZYT21} among others.

In this work, we use model-based methods for filtering toxic and pornographic content in addition to keyword and domain blocking, and we mask PII using regular expressions; see \ref{subsec:filter} for details.

\subsubsection{Data Quality}
Existing pre-training dataset processing pipelines incorporate various methods aimed at data quality. We categorize these into three main aspects: format accuracy, data redundancy, and content quality.

Format correctness primarily focuses on whether the data meets the expected format, such as whether it contains HTML tags, uncommon characters, or whether punctuation is used correctly. In existing research, this is mainly achieved through manual rules targeting the format or heuristic rules based on statistics. For example, clearing HTML tags or JavaScript code, normalizing special characters in the text, and counting the number and proportion of specific character sets, etc. Currently, almost all related studies include this step in their data processing workflow. After referring to the heuristic rules of Gopher and C4, we added some rules based on a deeper observation of the data. For detailed information, please refer to \ref{subsec:rule}.

``Duplication'' refers to the presence of repeated content in the data. Numerous studies have demonstrated that duplicate data not only wastes computational resources but can also negatively impact model performance. Current deduplication methods mainly fall into three categories: URL-based matching, string-based matching, and fuzzy matching.

URL-based matching methods primarily address duplicate data captured in Common Crawl, such as those employed by RefinedWeb and Dolma. String-based matching methods are used to detect duplicate text segments; for instance, RefinedWeb uses a global suffix tree to find duplicate segments, while Redpajama detects duplicates by calculating and comparing the hash values of each line in every document.

Fuzzy matching methods are mainly used to detect duplication across entire documents, with Locality-Sensitive Hashing (LSH) based methods being the most widely used. For example, Gopher, RefinedWeb, Slimpajama, Wudao, ChineseWebText, and Dolma all use Bloom Filters for deduplication operations. In this study, we chose to use an LSH-based deduplication method, for more details please refer to \ref{subsec:dedup}.

Content quality refers to whether the information in the dataset can effectively enhance the performance of the model. Typically, we intuitively measure its quality by training and evaluating a model of a certain scale on the dataset. However, this method requires substantial computational resources and is difficult to directly correspond with data filtering strategies. 

In existing research, the primary approach is to use known high-quality datasets as positive samples, then train language models (LM) or classifiers for scoring and filtering. For instance, in Redpajama's work, they used Wikipedia and random Common Crawl data as positive and negative samples respectively, trained a classifier, and used the classification probability of each document as a measure of its quality. Additionally, the Roots project trained a Ken-LM model using Wikipedia data, and used the perplexity of each document as a measure of its quality.

While using high-quality datasets as positive samples reduces additional investment, it lacks intuitive means for data quality assessment and is difficult to adjust specifically. To address the two widespread issues in Common Crawl: Advertising Content and Non-Fluent Content, we adopted a quality filtering method based on classification models in this study. For more details, please refer to \ref{subsec:quality-filter}.
\section{Method}
\begin{figure}[htbp]
    \centerline{
    \centering
    \includegraphics[width=\linewidth]{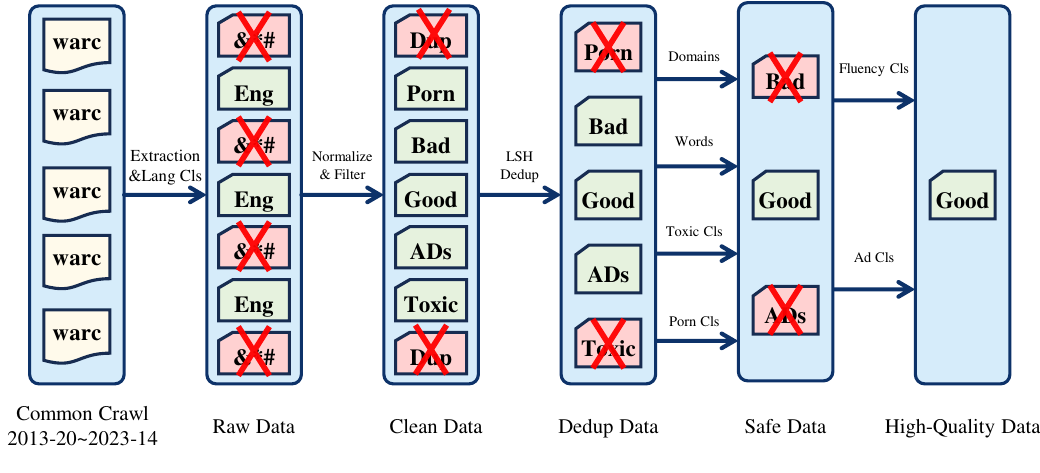}
    }
    \caption{Data Processing Pipeline. ``Warc'' represents the WARC format data from Common Crawl, ``Eng'' represents normal English data, ``\&*\#'' represents data with formatting errors, ``Dup'' represents duplicate data, ``Porn'' and ``Toxic'' represent pornographic and toxic data, respectively. ``ADs'' and ``Bad'' represent advertisements and low-quality data, respectively, while ``Good'' represents high-quality data. The red cross indicates data discarded at that stage.}
    \label{fig:pipline}
\end{figure}

\subsection{Processing Pipeline}
The data processing pipeline adopted in this study is shown in Figure \ref{fig:pipline}. Initially, we extract text from the WARC format data of Common Crawl, yielding the \textbf{Raw Data}. Subsequently, heuristic rules are applied to filter the \textbf{Raw Data}, generating the \textbf{Clean Data}. We then employ a deduplication method based on Locality-Sensitive Hashing (LSH) to process the \textbf{Clean Data}, resulting in the \textbf{Dedup Data}. Following this, we use filtering methods based on keywords and domain lists, as well as harmful content classifiers and obscene content classifiers based on Bert, to filter the \textbf{Dedup Data}, producing the \textbf{Safe Data}. Finally, we apply a Bert-based advertisement classifier and fluency classifier to further filter the \textbf{Safe Data}, obtaining the \textbf{High-Quality Data}.

\subsection{Data Extraction}
Each Common Crawl dump contains the original content of the crawled web pages, such as HTML, CSS, JS, and URLs of multimedia resources. To obtain relatively clean and coherent training corpora, we need to extract text information from these raw data. Additionally, at this stage, we can also perform language classification.
We employed the trafilatura method\cite{DBLP:conf/acl/Barbaresi21} for text extraction. Subsequently, we utilized the pycld2 library to classify the language of the extracted text. As this study is solely concerned with English data, we retained only the data marked as ``en'', discarding data in other languages.

\subsection{Heuristic Rule Filtering}
\label{subsec:rule}
Given the prevalence of parsing errors, formatting issues, and non-natural text in the extracted data, the typical solution is to design heuristic rules to modify and filter these data. Building on the heuristic rules established by Gopher\cite{DBLP:journals/corr/abs-2112-11446} and C4\cite{DBLP:conf/emnlp/DodgeSMAIGM021}, we have added some new rules based on our in-depth observation of the data. Firstly, we implement a set of rules that aim to normalize and simplify the content of the data, including character normalization, effective line extraction, line content modification, and line content removal. After applying these rules to truncate and modify the documents, we execute a series of heuristic rules to filter the entire document. For example, the ratio of letters in the text is too low, there are too many spaces and line breaks, there is a lack of punctuation, or too long words, etc. After the document has been intercepted, modified, and filtered according to the above rules, we obtained the \textbf{Clean Data}.

\subsection{Deduplication with MinHash-LSH}
\label{subsec:dedup}
Common Crawl contains a large amount of duplicate data, including reprinted articles, templated content, and repeatedly crawled information. This type of duplicate data can negatively impact the training effectiveness of models. To eliminate these duplicates, we employed a deduplication algorithm based on MinHash\cite{DBLP:conf/sequences/Broder97}.

Specifically, we utilized the datasketch library\cite{ericzhu2017290602} for computation, applying 128 hash functions on the 5-gram of the documents to construct signatures, and setting a certain threshold to identify highly similar data as duplicates. During the processing, we prioritized retaining the most recent data, meaning that Common Crawl dumps with larger numbers were considered first. After the deduplication process using LSH (Locality-Sensitive Hashing), we obtained the Dedup data.

\subsection{Safety Filtering}
\label{subsec:filter}
The web data contains a vast amount of unsafe content, including toxic, pornographic, and personally identifiable information (PII). To prevent pre-training models from learning undesirable behaviors from these contents, we initially employed a block domain list, block word list, toxicity classifier, and pornography classifier to tag and uniformly eliminate such data. Subsequently, we used regular expressions to mask any potential PII present in the data.

\subsubsection{Block Domain/Word List}
We created a blocklist of approximately 13 million unsafe domain names, along with a blocklist of 36,289 unsafe words. We employed domain matching and text matching methods to label all the data. Considering that unsafe words could potentially lead to excessive data deletion, we strived to minimize words that might cause such deletions.

To compensate for the limitations of the conservative block word list, we employed Bert-based Toxicity and Pornography classifiers to filter the data. The following sections will provide a detailed introduction to the preparation of training data and the training process for these two models, as well as report on their performance.

\subsubsection{Toxicity Classification}
To address the potential harmful content on the Internet, we utilized the training and test sets provided by the Toxic Comment Classification Challenge on the Kaggle platform\cite{jigsawtoxic} to fine-tune the Bert model for toxic data filtering. This dataset includes labels for six categories: toxic, severe\_toxic, obscene, threat, insult, and identity\_hate. We simplified these labels into a binary classification of "toxic" and "non-toxic", based on whether any of the original six attributes were present. In this way, we obtained a binary classification dataset. During this process, we did not alter the original division of the training and test sets in the dataset and fine-tuned the Bert model.

\subsubsection{Pornography Classification}
We utilized a publicly accessible list of pornographic website domains to extract a sample of 100,000 entries from our data. An equal number of non-pornographic samples were also extracted. These samples were proportionally drawn from various data dumps. To enhance the accuracy of our annotations, we employed an API for further annotation. This process resulted in a refined dataset of 60,000 items. We then selected a subset of this data for manual annotation as the test set, and used the remaining data to train and fine-tune our model.

\subsubsection{Personal Identifiable Information Masking}
To safeguard user privacy, we employed regular expressions to mask potential Personally Identifiable Information (PII) present in the data. The types of PII that we masked include: IP addresses, passport numbers, GPS location data, personal email addresses, personal phone numbers, identification card numbers, bank account numbers, and physical addresses. We used specific patterns for each type of PII, identified their positions in the text, and replaced them with a generic placeholder.

In summary, we employed four modules for data safety filtering, including: block domain list, block word list, toxicity classifier, and pornography classifier. Additionally, we utilized regular expressions to mask any potential personally identifiable information (PII). Ultimately, all data marked for discard were filtered out, with the remaining data considered as \textbf{Safe Data}.

\subsection{Data Quality Filtering}
\label{subsec:quality-filter}
After the initial filtering, we noticed that the data still included a significant amount of promotional content and poorly written text. To address this, we used a Bert-based classifier to further filter the data based on these two aspects.

\subsubsection{Advertisement Classification}
In dealing with advertisement attributes, we established a detailed manual annotation guide, requiring annotators to categorize the data into advertisements or not advertisements. thereby constructing a binary classification dataset, and divided into training and test sets.

\subsubsection{Fluency Classification}
For the attribute of fluency, we constructed a more comprehensive manual annotation questionnaire, requiring annotators to evaluate the data based on several dimensions including consistency, noise level, information value, and grammar correctness. We assign an overall tag(fluent or non-fluent) to each piece of data based on scores over all dimensions, thereby constructing a binary classification dataset, and divided into training and test sets.

We fine-tuned the BERT for sequence classification model using both of the aforementioned datasets. Subsequently, we utilized these two models to compute scores across the entirety of the \textbf{Safe Data}, and from this, we selected 1T tokens worth of data as \textbf{High-Quality Data}.

\subsection{Data Quality Evaluation}
After applying our data processing pipeline, we obtain a \textbf{High-Quality Data}. To ensure the quality of this data, we have implemented a comprehensive evaluation and feedback mechanism, taking into account the challenges posed by the large volume and diverse nature of the Common Crawl dataset.

In line with industry standards and best practices, we assess the dataset using a variety of quality evaluation metrics. These metrics cover several aspects such as effectiveness, completeness, understandability, similarity, fluency, relevance, and security. Each aspect includes multiple secondary metrics to provide a comprehensive evaluation.

To conduct a comprehensive, accurate, and efficient data quality evaluation, we employ three distinct methodologies: Human Evaluation, Quality Signal Evaluation, and Model Evaluation. These methods help us identify and measure the severity of data quality issues, and refine our evaluation and processing methods based on feedback from downstream model evaluations. 

\begin{figure}[htbp]
    \centerline{
    \centering
    \includegraphics[width=\linewidth]{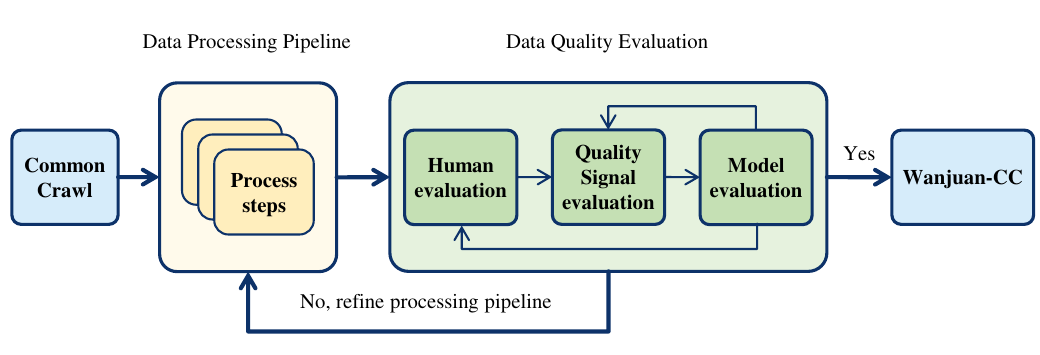}
    }
    \caption{Data Quality Evaluation Workflow}
    \label{fig:quality-eval}
\end{figure}

The current Data Quality Evaluation employs three distinct methods. Should all evaluations satisfy the quality requirement threshold, the dataset is deemed to be of high quality. Conversely, if this standard is not met, we proceed to optimize the data process and re-evaluate until the data quality threshold requirements are fulfilled. The comprehensive workflow of the Data Quality Evaluation process is depicted in Figure \ref{fig:quality-eval}.

\section{Result}
In order to more comprehensively illustrate the production process and data quality of WanJuan-CC, we have divided our work and results into four sections: document retention rate, distribution of statistical indicators across documents, data toxicity detection, and model evaluation.

Initially, we enumerated the number of documents at each stage within the data processing pipeline, computed both the relative removal rate and absolute retention rate, and depicted these in Figure \ref{fig:retention}. Subsequently, we calculated the distribution of various metrics within the dataset and represented these in Figure \ref{fig:statistics}. Following this, we employed the Perspective API to perform a safety assessment on the dataset, with AUC values presented in Table \ref{tab:toxicity_auc}. Lastly, we utilized an autoregressive decoder-only Transformer model to conduct experiments on diverse validation sets, with the experimental outcomes displayed in Table \ref{tab:ppl}.

\subsection{Removal Rate for Different Stages}

\begin{figure}[!htbp]
    \centering
    \includegraphics[width=\textwidth]{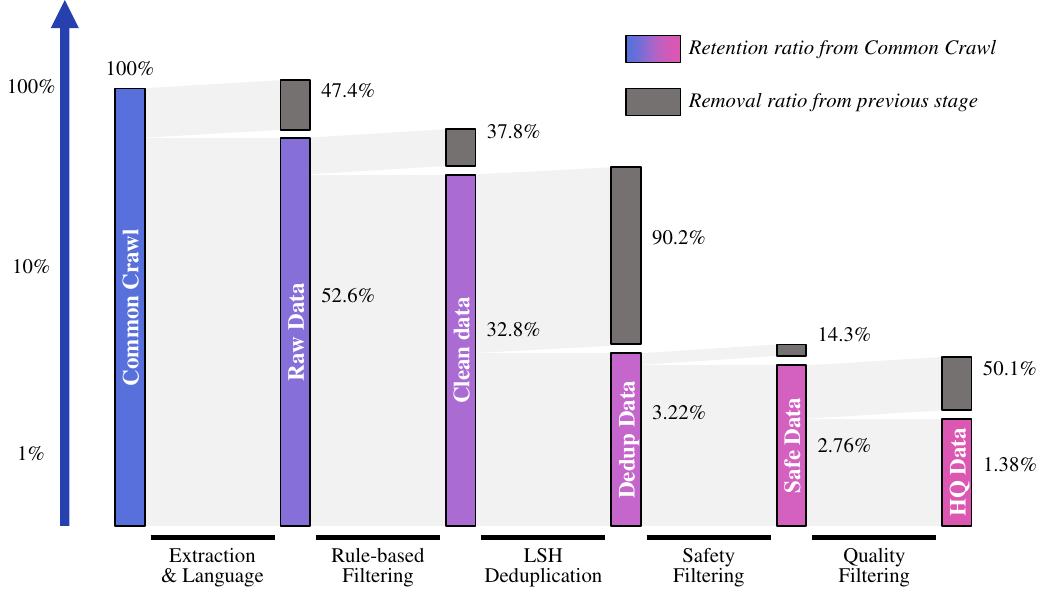}
    \caption{Documents Retention Rate and Removal Rate for Different Stages}
    \label{fig:retention}
\end{figure}

We quantified the document count at each phase of our data processing pipeline, encompassing the Common Crawl web pages, Raw data, Clean data, Deduplicated data, Safe data, and High-Quality data. We computed both the elimination rate for each phase in relation to its predecessor and the retention rate with respect to the total number of Common Crawl web pages. These metrics are graphically represented in Figure \ref{fig:retention}. It should be noted that a logarithmic scale was employed for the axes.

Initially, we extracted pertinent text from the HTML data embedded in the WARC format files of Common Crawl, identifying the English content therein. This process retained 52.6\% of the total documents. Following this, a heuristic rule-based data cleansing step was implemented, which filtered out approximately 37.8\% of the data. This step primarily served to eliminate data that was clearly not part of standard text or exhibited severe formatting errors.

In the deduplication phase, our elimination rate markedly diverged from that of other studies. Our LSH deduplication approach expunged 90.2\% of the data. Conversely, RefinedWeb's combined strategy of fuzzy deduplication and exact deduplication discarded a total of 50\% of the data. The presence of duplicates in pre-training data can substantially impair model performance, hence our adoption of a more rigorous deduplication plan.

In the process of safe content filtering, a combination of sensitive word detection, domain blocking, pornographic classification, and toxicity classification eliminated 14.3\% of the data. Subsequently, during the quality filtering phase, we discarded approximately 50\% of the remaining data. The final high-quality dataset constituted 1.48\% of the original data volume.

\begin{figure}[!htbp]
    \centering
    \includegraphics[width=\textwidth]{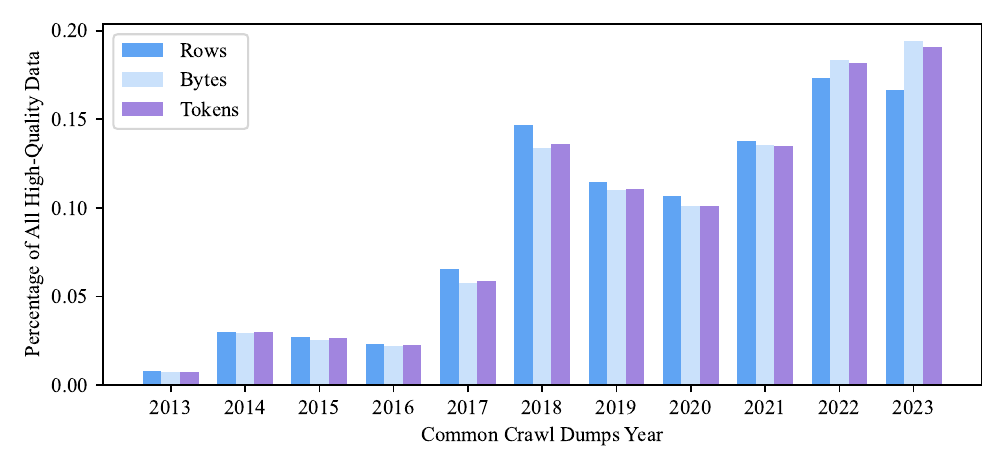}
    \caption{Proportion of High-Quality Data from Different Common Crawl Dump Years}
    \label{fig:data_volume}
\end{figure}

We also calculated the proportion of high-quality data originating from different years of Common Crawl dump, as depicted in Figure \ref{fig:data_volume}. As seen from the figure, the number of documents, bytes, and tokens in high-quality data have all shown an increasing trend over time. This could be attributed to two main factors. On one hand, it is due to the continuous growth of the data volume in Common Crawl. On the other hand, it is related to our retention strategy implemented during the LSH deduplication step, which prioritizes the preservation of the most recent data.

\subsection{Dataset Statistics}
\begin{figure}
    \centering
    \begin{subfigure}{.3\textwidth}
        \centering
        \includegraphics[width=\linewidth]{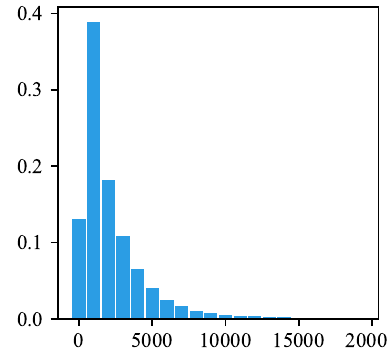}
        \caption{Content Length}
    \end{subfigure}%
    \begin{subfigure}{.3\textwidth}
        \centering
        \includegraphics[width=\linewidth]{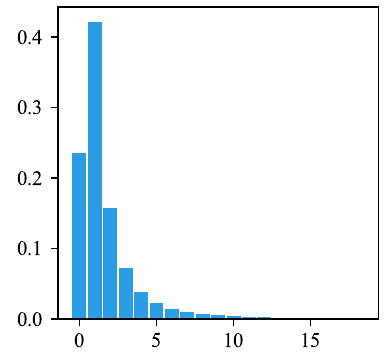}
        \caption{Line Number}
    \end{subfigure}
    \begin{subfigure}{.3\textwidth}
        \centering
        \includegraphics[width=\linewidth]{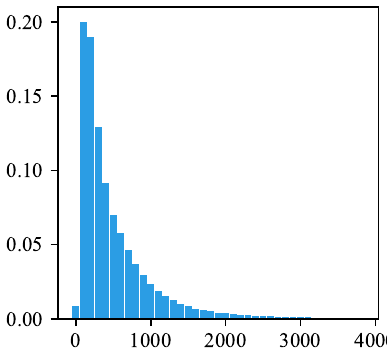}
        \caption{Token Length}
    \end{subfigure}
    
    \begin{subfigure}{.3\textwidth}
        \centering
        \includegraphics[width=\linewidth]{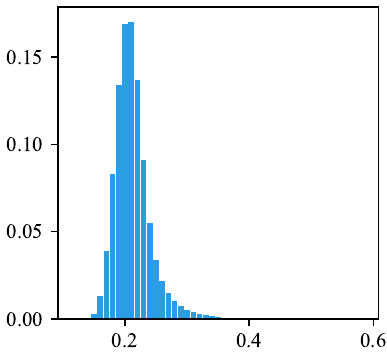}
        \caption{Non-alpha Fraction}
    \end{subfigure}%
    \begin{subfigure}{.3\textwidth}
        \centering
        \includegraphics[width=\linewidth]{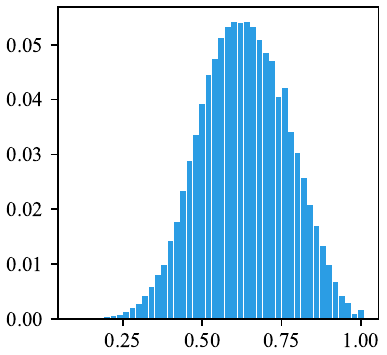}
        \caption{Unique Words Fraction}
    \end{subfigure}
    \begin{subfigure}{.3\textwidth}
        \centering
        \includegraphics[width=\linewidth]{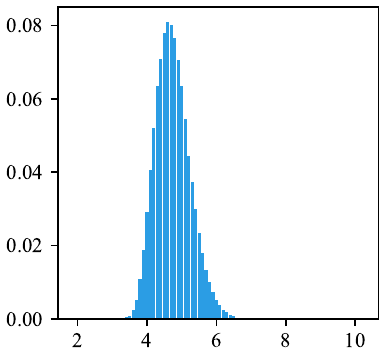}
        \caption{Mean Word Length}
    \end{subfigure}
    
    \begin{subfigure}{.3\textwidth}
        \centering
        \includegraphics[width=\linewidth]{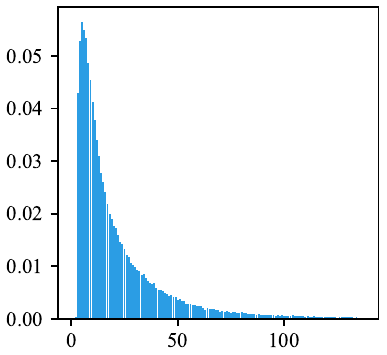}
        \caption{Sentence Number}
    \end{subfigure}%
    \begin{subfigure}{.3\textwidth}
        \centering
        \includegraphics[width=\linewidth]{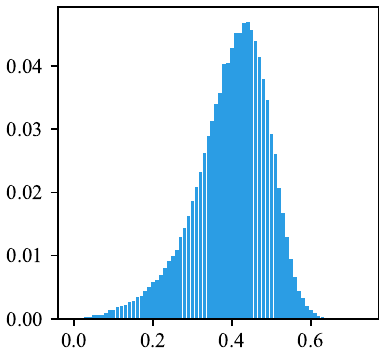}
        \caption{Stop Word Fraction}
    \end{subfigure}
    \begin{subfigure}{.3\textwidth}
        \centering
        \includegraphics[width=\linewidth]{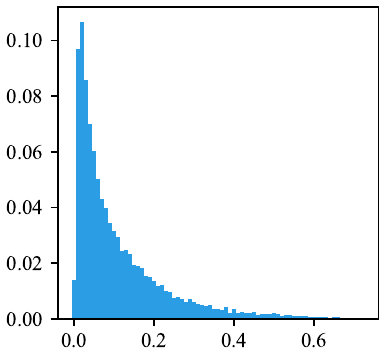}
        \caption{Symbol to Word Ratio}
    \end{subfigure}
    \caption{Percentage Statistics for Different Metrics on WanJuan-CC. To illustrate the main distribution area, the statistical range of some metrics has been truncated due to the presence of long-tail distributions.}
    \label{fig:statistics}
\end{figure}

In order to streamline the assessment of the dataset's quality, we drew upon several data quality metrics from Redpajama-v2 and applied them to WanJuan-CC. We quantified various aspects of the dataset including document length, line count, token length, percentage of non-alphabetic characters, proportion of unique words, average word length, sentence count, stop-word ratio, and symbol-to-word ratio. The distribution for each metric is visualized in Figure \ref{fig:statistics}.

By quantifying and illustrating the distribution of these metrics on WanJuan-CC, we enable users to gain a comprehensive understanding of the data's various characteristics. This empowers users to refine the data further in accordance with their specific requirements.

\subsection{Data Safety Metrics}
We independently drew a random sample of 100K data points from each of the following datasets: our own, Redpajama, and RefinedWeb. We utilized the Perspective API to compute scores across seven dimensions. Let's denote the test dataset of any given dataset as $D=\{d_j|1\le j\le N\}$, where for a specific dimension $m$, the score of data point $d_j$ is represented by $c_m(d_j)$. Given a threshold $t_i$, the percentage of data exceeding $t_i$ is calculated as $p_m(t_i)=100 \times \frac{|\{d_j|c_m(d_j)>t_i\}|}{N}$. 

By generating the ``$t$-$p_m(t)$'' curves and computing the AUC values for each curve using Equation \ref{eq:toxic_auc}, we can intuitively assess the level of unsafety across different dimensions. Detailed results are presented in Table \ref{tab:toxicity_auc}. The table reveals that our dataset's AUC values are lower than those of both Redpajama and RefinedWeb across all dimensions, suggesting a higher degree of safety in our dataset.
\begin{equation}
    \label{eq:toxic_auc}
    AUC_m = \sum_{i=1}^{L-1} \frac{p_m(t_i)+p_m(t_{i+1})}{2} \times (t_{i+1}-t_i)
\end{equation}

\begin{table}[ht]
    \centering
    \caption{Area Under Curve of Toxicity Metrics by Perspective API for Different Datasets}
    \begin{tabular}{cccc}
    \toprule
     & WanJuan-CC & Redpajama & RefinedWeb \\
    \midrule
    Toxicity & \textbf{3.45} & 4.02 & 4.48 \\
    Severe Toxicity & \textbf{0.21} & 0.31 & 0.46 \\
    Identity Attack & \textbf{0.83} & 1.35 & 1.14 \\
    Insult & \textbf{1.83} & 2.16 & 2.39 \\
    Profanity & \textbf{1.88} & 2.30 & 2.88 \\
    Threat & \textbf{1.02} & 1.31 & 1.19 \\
    Sexually Explicit & \textbf{1.28} & 2.19 & 2.42 \\
    \bottomrule
    \end{tabular}
    \label{tab:toxicity_auc}
\end{table}

\subsection{Data Utility Metrics}
\label{subsec:data_utility_metric}
To further ascertain the quality of the data, we utilized the identical autoregressive decoder-only Transformer model for training from scratch with both WanJuan-CC and RefinedWeb datasets. Experiments were conducted at 1B and 3B parameter levels, where the 1B model processed 100B tokens and the 3B model processed 200B tokens.

We employed the average PPL on the validation dataset as the evaluation metric for the 1B model. This approach is due to the difficulty in observing changes in downstream task metrics with models of smaller parameter sizes, thus making the use of PPL metric on the validation set a more effective measure of training outcomes for smaller models. This perspective aligns with that expressed in Paper D4 \cite{DBLP:journals/corr/abs-2308-12284}. We incorporated three subsets from Pile \cite{DBLP:journals/corr/abs-2101-00027} - ``pile-books3'', ``pile-openwebtext2'', ``pile-wikipedia-en'' - and one subset from Tiny-stroys \cite{DBLP:journals/corr/abs-2305-07759} - ``tiny-storys'' - as our four validation sets. The experimental results are depicted in Table \ref{tab:ppl}. As can be discerned from the table, the PPL of WanJuan-CC on pile-books3, pile-openwebtext2, and pile-wikipedia-en is marginally lower than that of RefinedWeb, while its PPL on tiny-storys is markedly lower. These findings suggest that the data quality of WanJuan-CC slightly surpasses that of RefinedWeb, particularly on validation sets like tiny-storys that demand higher language fluency.

\begin{table}[ht]
    \centering
    \caption{Perplexity of Different Models on Different Validation Sets}
    \begin{tabular}{ccc}
    \toprule
    Eval Datasets  & WanJuan-CC & RefinedWeb \\
    \midrule
    pile-books3 & \textbf{12.34} & 12.56 \\
    pile-openwebtext2 & \textbf{11.86} & 11.96 \\
    pile-wikipedia-en & \textbf{8.68} & 8.81 \\
    tiny-storys & \textbf{5.78} & 6.15 \\
    \bottomrule
    \end{tabular}
    \label{tab:ppl}
\end{table}

For models with a larger number of parameters, the accuracy of downstream tasks can serve as a more effective indicator of data quality. We have chosen to evaluate three major categories of downstream tasks: English text completion (LAMBADA, StoryCloze), English general capability (SuperGLUE), and English commonsense question answering (HellaSwag, PIQA, WinoGrande), amounting to six downstream tasks in total. The results of these experiments are displayed in Table \ref{tab:downstream}.

\begin{table}[ht]
    \centering
    \caption{Accuracy of Different Models on Different Downstream Tasks}
    \begin{tabular}{ccc}
    \toprule
     Task & RefinedWeb & WanJuan-CC \\
    \midrule
    LAMBADA    & 58.65 & \textbf{62.22} \\
    StoryCloze & 70.44 & \textbf{71.83} \\
    SuperGLUE  & 41.86 & \textbf{44.46} \\
    HellaSwag  & \textbf{60.71} & 58.12 \\
    PIQA       & \textbf{74.92} & 74.21 \\
    WinoGrande & 55.33 & \textbf{57.54} \\
    \bottomrule
    \end{tabular}
    \label{tab:downstream}
\end{table}

Experimental results demonstrate that WanJuan-CC significantly enhances performance in English text completion and general English proficiency, albeit with a decline in HellaSwag and a minor decrease on PIQA for English commonsense question answering. Conversely, there is a substantial improvement on WinoGrande. In summary, WanJuan-CC surpasses RefinedWeb in terms of performance on downstream tasks.

\section{Conclusion}
In this paper, we presented WanJuan-CC, a safe and high-quality open-sourced English webtext dataset derived from Common Crawl data. We designed and implemented a comprehensive process for handling Common Crawl data, which includes data extraction, heuristic rule filtering, fuzzy deduplication, content safety filtering, and data quality filtering. This process allowed us to extract approximately 68 billion original English documents from 90 dumps since 2013, yielding 2.22T Tokens of safe data and 1.0T Tokens of high-quality data.The additional steps of advertisement and fluency classification on dataset were key to ensuring high data quality.

The safety of our dataset was validated through rigorous testing. We used the Perspective API to assess the level of unsafety across different dimensions, and our dataset's AUC values were found to be lower than those of other datasets across all dimensions, indicating a higher degree of safety in our dataset. Furthermore, the utility of our dataset was demonstrated through perplexity (PPL) on validation sets and accuracy on downstream tasks. The PPL of WanJuan-CC on various validation sets was found to be competitive, particularly on sets like tiny-storys that demand higher language fluency. For larger models, the accuracy of downstream tasks served as an effective indicator of data quality. Experimental results showed that WanJuan-CC significantly enhances performance in English text completion and general English proficiency tasks.

In conclusion, WanJuan-CC represents a significant contribution to the field of large-scale language model training. It provides a safe, high-quality, and open-source dataset for researchers and practitioners. Future work could focus on further refining the data processing pipeline to improve data quality and safety, and exploring the application of this dataset in more diverse NLP tasks.
\newpage

\bibliographystyle{unsrt}
\bibliography{wanjuan_v2.bib}

\end{document}